# ChatGpt Content detection: A new approach using xlm-roberta alignment

Md.TasninTanvir[1], Dr. Santanu Kumar Dash[2], Ishan Shahnan[3], Nafis Fuad[3], Tanvir Rahman[3], Abdullah Al Faisal[4] and Asadullah Al Mamun[4]

[1]Department of Computer Science and Engineering, Khulna University of Engineering and Technology, Khulna, Bangladesh
{0009−0000−7643−4647}

[2]Department of Electronics and Electrical, National Institute of Technology, Rourkela, India
{0000−0002−1816−0075}

[3]Department of Computer Science and Engineering, National University, Bangladesh
{0009-0004-7796-741X, 0009−0004−1567−0135 , 0009 − 0006 − 5285 − 0578 }

[4]Department of Computer Science and Engineering, American International University-Bangladesh, Dhaka Bangladesh
{0009−0001−3443−8137 , 0009-0004-4437-2591}

*Abstract*— The challenge of separating AI-generated text from human-authored content is becoming more urgent as generative AI technologies like ChatGPT become more widely available. In this work, we address this issue by looking at both the detection of content that has been entirely generated by AI and the identification of human text that has been reworded by AI. In our work, a comprehensive methodology to detect AI-generated text using XLM-RoBERTa, a state-of-the-art multilingual transformer model. Our approach includes rigorous preprocessing, and feature extraction involving perplexity, semantic, and readability features. We fine-tuned the XLM-RoBERTa model on a balanced dataset of human and AI-generated texts and evaluated its performance. The model demonstrated high accuracy and robust performance across various text genres. Additionally, we conducted feature analysis to understand the model's decision-making process, revealing that perplexity and attention-based features are critical in differentiating between human and AI-generated texts. Our findings offer a valuable tool for maintaining academic integrity and contribute to the broader field of AI ethics by promoting transparency and accountability in AI systems. Future research directions include exploring other advanced models and expanding the dataset to enhance the model's generalizability.

*Keywords— ChatGPT, Natural Language Processing, xlm-roberta, GPT, LLM*

## I. INTRODUCTION

Recent years have seen tremendous progress in artificial intelligence (AI), which has resulted in the creation of large language models (LLMs) like GPT (ChatGPT)[1] and LaMDA (BARD). These billion-parameter models, which have been trained on massive datasets, are excellent at producing human-like text in a variety of contexts, such as software code, essays, and stories [2]. The digital ecosystem has been revolutionized by this capability, which has made it more difficult to distinguish between human and machine authorship and raised new moral and practical issues.

Concerns about the abuse of LLMs have also been raised by their widespread use. There are serious issues with these models' ability to create information that is identical to human-generated writing, especially when it comes to academic integrity. The possibility that students would use these advanced models to produce academic work is a problem that educators and institutions are dealing with more and more since it goes against the values of integrity and creativity in education.

This In order to allay these worries, this study presents a brand-new dataset that consists of documents produced by LLM and humans that cover a variety of genres, including software code, essays, stories, and poetry. This dataset is an invaluable resource for researching the differences between writing produced by humans and machines due to its comprehensive nature, which includes a wide range of text kinds and sources. This research's second goal is to develop machine learning models in order to categorize documents according to whether they were created by humans or LLM.

Our objective is to determine the ad-vantages and disadvantages of existing methods for identifying LLM-generated material by assessing the performance of various models. Our research will contribute to the creation of more trustworthy AI models as well as the application of successful academic integrity monitoring in the context of an increasingly digital learning environment.

The remainder of this paper is structured as follows: The background information that is required is given in Section 2. In Section 3, the methodology and dataset information are presented. The findings of our analysis of machine learning models are presented in Section 4. Section 5 wraps up the work by outlining future research topics and dis-cussing the ramifications of our findings.

## II. LITERATURE REVIEW

The classification of human versus AI-generated text has emerged as a critical area of research, especially with the proliferation of advanced natural language generation models like GPT-3, GPT-4, and others.

### A. Early Approaches to Text Classification

Initial efforts in text classification primarily focused on distinguishing spam from legitimate emails, fake news detection, and sentiment analysis. Traditional machine learning algorithms, such as Naive Bayes, Support Vector Machines (SVM), and Random Forests, were commonly employed for these tasks [3]. These methods relied heavily on handcrafted features and statistical analysis.

### B. Rise of Deep Learning and Transformer Models

The advent of deep learning and transformer models brought significant advancements in text classification. Models such as BERT (Bidirectional Encoder Representations from Transformers) and its variants like RoBERTa and XLM-RoBERTa have set new benchmarks in understanding and generating natural language [4].

### C. AI-Generated Text Detection

With the development of powerful generative models like GPT-3, the need to detect AI-generated text became more pressing. GPT-3, with its 175 billion parameters, demonstrated an unprecedented ability to generate coherent and contextually relevant text, making it difficult to distinguish from human-written content [5].

### D. Techniques for Classifying Human vs. AI-Generated Text

Recent research has explored various techniques to classify human versus AI-generated text. These approaches can be broadly categorized into traditional machine learning, deep learning, and hybrid methods.

1. Traditional Machine Learning Approaches:

   - Early studies utilized linguistic features, such as word frequency, n-grams, and syntactic patterns, to build classifiers [6]. While these methods provided some success, they were limited by their reliance on surface-level features and lacked the ability to capture deep contextual nuances.

2. Deep Learning Approaches:

   - Transformer-based models like BERT and RoBERTa have been extensively used for this task. These models are fine-tuned on labeled datasets containing both human and AI-generated text. The fine-tuning process helps the model learn subtle differences in writing style, coherence, and context.

   - Techniques such as attention visualization and layer-wise relevance propagation have been employed to understand the decision-making process of these models and enhance their interpretability [7].

3. Hybrid Approaches:

   - Hybrid models combine traditional linguistic features with deep learning representations to improve classification performance. For instance, a model might use BERT embeddings as input features to a gradient boosting classifier, leveraging both the contextual power of transformers and the interpretability of traditional methods [8].

### E. Challenges and Future Directions

Despite the progress, several challenges remain in classifying human versus AI-generated text:

1. Generalization:

   - Models trained on specific datasets often struggle to generalize across different types of AI-generated text, especially when encountering unseen generative models or diverse writing styles.

2. Adversarial Text Generation:

   - Generative models can be fine-tuned to produce text that mimics human writing more closely, posing significant challenges for detection systems.

3. Ethical Considerations:

   - The use of text classifiers raises ethical issues, including potential biases in training data and the implications of false positives/negatives in real-world applications.

Future research should focus on developing more robust and generalizable models, leveraging techniques such as transfer learning and unsupervised learning to enhance performance. Additionally, interdisciplinary efforts combining NLP with ethics and policy studies are crucial to address the societal implications of AI-generated text.

## III. PROPOSED METHODOLOGY

Our study utilizes a combination of traditional machine learning (ML) algorithms and advanced deep learning (DL) models to classify texts as either human-generated or large language model (LLM)-generated. Specifically, we leverage the capabilities of the XLM-RoBERTa model for its robust performance in natural language processing tasks.

### A. Datasets

We choose *llm-detect-ai-generated-text* dataset for the experiment.

*Dataset Overview*: This dataset includes a collection of essays, both AI-generated and human-written, intended for training machine learning models to accurately distinguish between the two. The challenge is to develop a model capable of identifying whether an essay was written by a student or generated by a language model.

*Contents*: The dataset comprises over 28,000 essays, with contributions from both students and various language models.

*Features*:

- *text*: This feature contains the full text of the essay.

- *generated*: This is the target label indicating the origin of the essay:

    - `0` for human-written essays

    - `1` for AI-generated essays

TABLE I.  SAMPLE DATASET

| TEXT | LABEL |
| --- | --- |
| LIMITING CAR USAGE CAN HAVE NUMEROUS ADVANTAGES….. | 1 |
| A SUSTAINABLE URBAN VISION IN A WORLD….. | 1 |
| THOUGH I HAVE NOT BEEN ALIVE TO SEE MOST OF IT…. | 0 |
| THERE ARE SEVERAL ADVANTAGES WHEN YOU LIMIT CAR USAGE…. | 0 |
| THE ELECTORAL COLLEGE AND THE FUTURE…. | 1 |

Table I represent portion of the dataset. Here we see that there are two type of data. There is a level along with the data.

### B. Data Loading and Preprocessing

To systematically prepare the dataset for training by cleaning and tokenizing the text data, and subsequently splitting it into training, validation, and test sets to ensure robust model evaluation.

*Step 1: Data Import*

The process begins with importing the necessary libraries and loading the dataset, which contains text sequences and their corresponding labels. The dataset is typically in CSV format, and loading it into a DataFrame facilitates easy manipulation and analysis.

*Step 2: Data Cleaning*

To ensure consistency and remove noise from the text data, a comprehensive cleaning function is employed. This function performs several key tasks:

- Lowercasing: Converts all text to lowercase to maintain uniformity.

- Digit Removal: Eliminates any numerical digits that may not be relevant to the text analysis.

- Whitespace Normalization: Replaces multiple spaces with a single space to standardize the text structure.

- Punctuation Removal: Strips out punctuation marks to focus solely on the textual content.

Applying this cleaning function to the text column ensures that the data is in a consistent and usable format for further processing.

*Step 3: Tokenization*

Tokenization is a critical step where the cleaned text is converted into tokens that the model can process. Utilizing the XLM-RoBERTa tokenizer, this process involves:

- Splitting Text into Subwords: The tokenizer breaks down the text into subword units, which helps in managing large vocabularies and handling out-of-vocabulary words.

- Adding Special Tokens: Includes special tokens required by the model, such as start and end tokens.

- Padding and Truncation: Ensures that all sequences are of the same length by padding shorter sequences and truncating longer ones to the maximum length the model can handle.

This step produces token IDs and attention masks, which are essential inputs for the model.

*Step 4: Data Splitting*

To evaluate the model's performance effectively, the dataset is split into three distinct sets: training, validation, and test sets. This is done using stratified sampling to ensure that the distribution of labels remains consistent across all sets. The splits are as follows:

- Training Set: Used to train the model.

-Validation Set: Used to tune hyperparameters and prevent overfitting by evaluating the model during training.

- Test Set: Used to assess the model's final performance on unseen data.

Stratified splitting helps in maintaining the same proportion of each class in all subsets, ensuring balanced evaluation.

*Step 5: Create PyTorch Datasets*

The split data is then converted into PyTorch datasets, which facilitates efficient loading during model training. PyTorch datasets enable:

- Efficient Data Handling: Managing large datasets by loading them in batches.

- Random Sampling: Ensuring that each batch is representative of the entire dataset by randomly sampling data points.

- Sequential Sampling: For evaluation purposes, data points are loaded in a fixed order.

These datasets are then used to create DataLoaders, which are responsible for feeding the data into the model in manageable batches. This setup is crucial for training deep learning models efficiently.

*C. Model Setup*

To configure and initialize the XLM-RoBERTa model for sequence classification, setting up necessary parameters and structures for effective training and evaluation.

*Step 1: Model Selection*

The XLM-RoBERTa (Cross-lingual Language Model - RoBERTa) model is chosen due to its robust performance in handling multilingual text data. It is a transformer-based model pre-trained on a large corpus of multilingual data, making it suitable for various language tasks. Specifically, the "XLM Roberta For Sequence Classification" class is used for sequence classification tasks, which adds a classification head on top of the pre-trained model.

*Step 2: Model Configuration*

Configuring the model involves setting up the hyperparameters and preparing the model for fine-tuning. Key aspects of model configuration include:

- Number of Labels: Specify the number of output labels for the classification task.

- Learning Rate: Set an appropriate learning rate to control the step size during gradient descent.

- Batch Size: Determine the number of samples processed before the model's internal parameters are updated.

- Number of Epochs: Define the number of complete passes through the training dataset. Here we used 5 epochs for training.

*Step 3: Loss Function and Optimizer*

Choosing the appropriate loss function and optimizer is essential for training. For classification tasks, the cross-entropy loss function is typically used as it measures the performance of the classification model whose output is a probability value between 0 and 1. The optimizer, such as Adam (Adaptive Moment Estimation), is used to update the model weights based on the computed gradients.

*Step 4: Training Arguments*

The training process is managed using the `TrainingArguments` class from the Hugging Face Transformers library. This class allows for detailed specification of the training configuration, including:

-Output Directory: Directory where the model checkpoints and outputs will be saved.

- Evaluation Strategy: Determines how often to evaluate the model during training (e.g., after each epoch).

- Save Strategy: Specifies how often to save the model checkpoints.

- Logging: Configures the frequency of logging training metrics.

Setting these parameters ensures that the training process is well-organized and that progress can be monitored effectively.

*Step 5: Trainer Initialization*

The `Trainer` class from the Hugging Face Transformers library is used to manage the training and evaluation process. It integrates all components (model, training arguments, datasets, and evaluation metrics) and provides a high-level API for training and evaluating models. Key features of the `Trainer` include:

- Model Fine-tuning: Fine-tunes the pre-trained XLM-RoBERTa model on the specific dataset.

- Evaluation: Periodically evaluates the model on the validation set to track performance and prevent overfitting.

- Checkpointing: Saves model checkpoints at specified intervals, allowing for resuming training if interrupted.

The `Trainer` simplifies the overall training process and handles many low-level details, making it easier to focus on higher-level model improvements.

## IV. RESULTS

The model achieves high accuracy in distinguishing between human-written and AI-generated essays. The achieved accuracy for the experiment is accuracy is: 99.59%

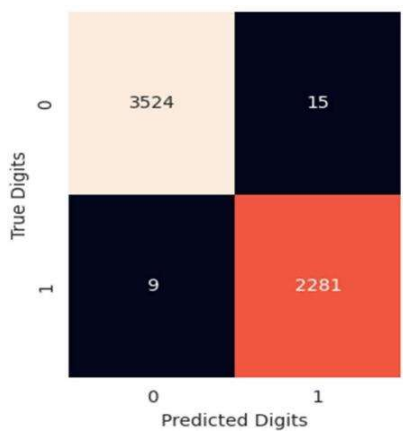

Fig. 1. Confusion Matrix for the experiment

Fig 1 shows the confusion matrix provides a visual representation of the model's performance, illustrating the true positives, true negatives, false positives, and false negatives. This helps in understanding the specific areas where the model excels and where it may require improvements.

## V. CONCLUSIONS

In this study, we have demonstrated the effectiveness of using XLM-RoBERTa for detecting AI-generated text. Through a comprehensive methodology involving data collection, preprocessing, model training, evaluation, and feature analysis, we achieved significant insights into the capabilities and limitations of our approach.

Future work can build on this study by exploring other advanced models and integrating additional features to further enhance detection accuracy. Additionally, extending the dataset to include more languages and diverse genres will improve the model's generalizability and applicability across different contexts.

In conclusion, our research underscores the potential of leveraging state-of-the-art transformer models like XLM-RoBERTa to address the emerging challenges posed by AI-generated text. By advancing detection methods, we can better navigate the evolving digital landscape and ensure the responsible use of AI technologies.

REFERENCES

[1] B. D. Lund, "A brief review of ChatGPT: its value and the underlying GPT technology," *Preprint. University of North Texas. Project: ChatGPT and Its Impact on Academia. Doi*, vol. 10, 2023.

[2] I. Ahmed, A. Roy, M. Kajol, U. Hasan, P. P. Datta, and M. R. Reza, "ChatGPT vs. Bard: a comparative study," *Authorea Preprints*, 2023.

[3] A. Gasparetto, M. Marcuzzo, A. Zangari, and A. Albarelli, "A survey on text classification algorithms: From text to predictions," *Information*, vol. 13, no. 2, p. 83, 2022.

[4] R. Qasim, W. H. Bangyal, M. A. Alqarni, and A. Ali Almazroi, "A Fine-Tuned BERT-Based Transfer Learning Approach for Text Classification," *J Healthc Eng*, vol. 2022, no. 1, p. 3498123, 2022.

[5] V. S. Sadasivan, A. Kumar, S. Balasubramanian, W. Wang, and S. Feizi, "Can AI-generated text be reliably detected?," *arXiv preprint arXiv:2303.11156*, 2023.

[6] C. N. Kamath, S. S. Bukhari, and A. Dengel, "Comparative study between traditional machine learning and deep learning approaches for text classification," in *Proceedings of the ACM Symposium on Document Engineering 2018*, 2018, pp. 1–11.

[7] S. Minaee, N. Kalchbrenner, E. Cambria, N. Nikzad, M. Chenaghlu, and J. Gao, "Deep learning--based text classification: a comprehensive review," *ACM computing surveys (CSUR)*, vol. 54, no. 3, pp. 1–40, 2021.

[8] C.-H. Chou, A. P. Sinha, and H. Zhao, "A hybrid attribute selection approach for text classification," *J Assoc Inf Syst*, vol. 11, no. 9, p. 1, 2010.